\documentclass[10pt,twocolumn]{article}

\usepackage[utf8]{inputenc}
\usepackage[T1]{fontenc}
\usepackage{lmodern}
\usepackage{geometry}
\usepackage{graphicx}
\usepackage{microtype}
\usepackage{amsmath,amssymb}
\usepackage{booktabs}
\usepackage{hyperref}
\usepackage[numbers]{natbib}

\usepackage{amsmath} 
\usepackage{bm}
\usepackage{subfig} 
\usepackage{placeins}      
\usepackage{multirow}
\usepackage{booktabs}
\usepackage{multirow}
\usepackage{pifont}
\usepackage{float}  
\usepackage{algorithm}
\usepackage{algpseudocode}  
\usepackage{amsmath}        
\usepackage[table]{xcolor}  
\usepackage{enumitem}
\title{\bfseries Detecting Hallucinations in Graph Retrieval-Augmented Generation via Attention Patterns and Semantic Alignment}

\author{
\begin{tabular}{c}
Shanghao Li\textsuperscript{1} \quad
Jinda Han\textsuperscript{2} \quad
Yibo Wang\textsuperscript{1} \quad
Yuanjie Zhu\textsuperscript{1} \\
Zihe Song\textsuperscript{1} \quad
Langzhou He\textsuperscript{1} \quad
Kenan Kamel A Alghythee\textsuperscript{1} \quad
Philip S. Yu\textsuperscript{1}
\\[6pt]
\textsuperscript{1}University of Illinois Chicago \\
\textsuperscript{2}University of Illinois Urbana-Champaign \\
\\[-2pt]
\texttt{sli261@uic.edu},
\texttt{psyu@uic.edu}
\end{tabular}
}

\date{}

\begin{document}
\maketitle
\begin{abstract}
Graph-based Retrieval-Augmented Generation (GraphRAG) enhances Large Language Models (LLMs) by incorporating external knowledge from linearized subgraphs retrieved from knowledge graphs. 
However, LLMs struggle to interpret the relational and topological information in these inputs, resulting in hallucinations that are inconsistent with the retrieved knowledge.
To analyze how LLMs attend to and retain structured knowledge during generation, we propose two lightweight interpretability metrics: Path Reliance Degree (PRD), which measures over-reliance on shortest-path triples, and Semantic Alignment Score (SAS), which assesses how well the model’s internal representations align with the retrieved knowledge.
Through empirical analysis on a knowledge-based QA task, we identify failure patterns associated with over-reliance on salient paths and weak semantic grounding, as indicated by high PRD and low SAS scores. We further develop a lightweight post-hoc hallucination detector, Graph Grounding and Alignment (GGA), which outperforms strong semantic and confidence-based baselines across AUC and F1.
By grounding hallucination analysis in mechanistic interpretability, our work offers insights into how structural limitations in LLMs contribute to hallucinations, informing the design of more reliable GraphRAG systems in the future.
\end{abstract}

\section{Introduction}
Large Language Models (LLMs) have shown strong performance in question answering and language generation. 
However, their reliance on parametric memory can lead to hallucinations, where the model produces fluent responses that contradict or lack support from factual knowledge \cite{huang2025survey}.
Retrieval-Augmented Generation (RAG) methods address this limitation by retrieving relevant external knowledge to ground LLM responses in factual context \cite{lewis2020retrieval,chen2024benchmarking}.
Graph-based RAG (GraphRAG) extends traditional RAG by retrieving structured subgraphs composed of entity–relation triples from a knowledge graph, rather than relying on unstructured text passages \cite{li2025towards,sun2025dyg}.
It improves reasoning by modeling entity relationships through structured subgraphs, making it more effective than RAG at handling multi-hop questions and aggregating information across noisy or distributed sources \cite{han2025rag, xiang2025use}.

\begin{figure}[h]
   \centering
   \includegraphics[width=\linewidth]{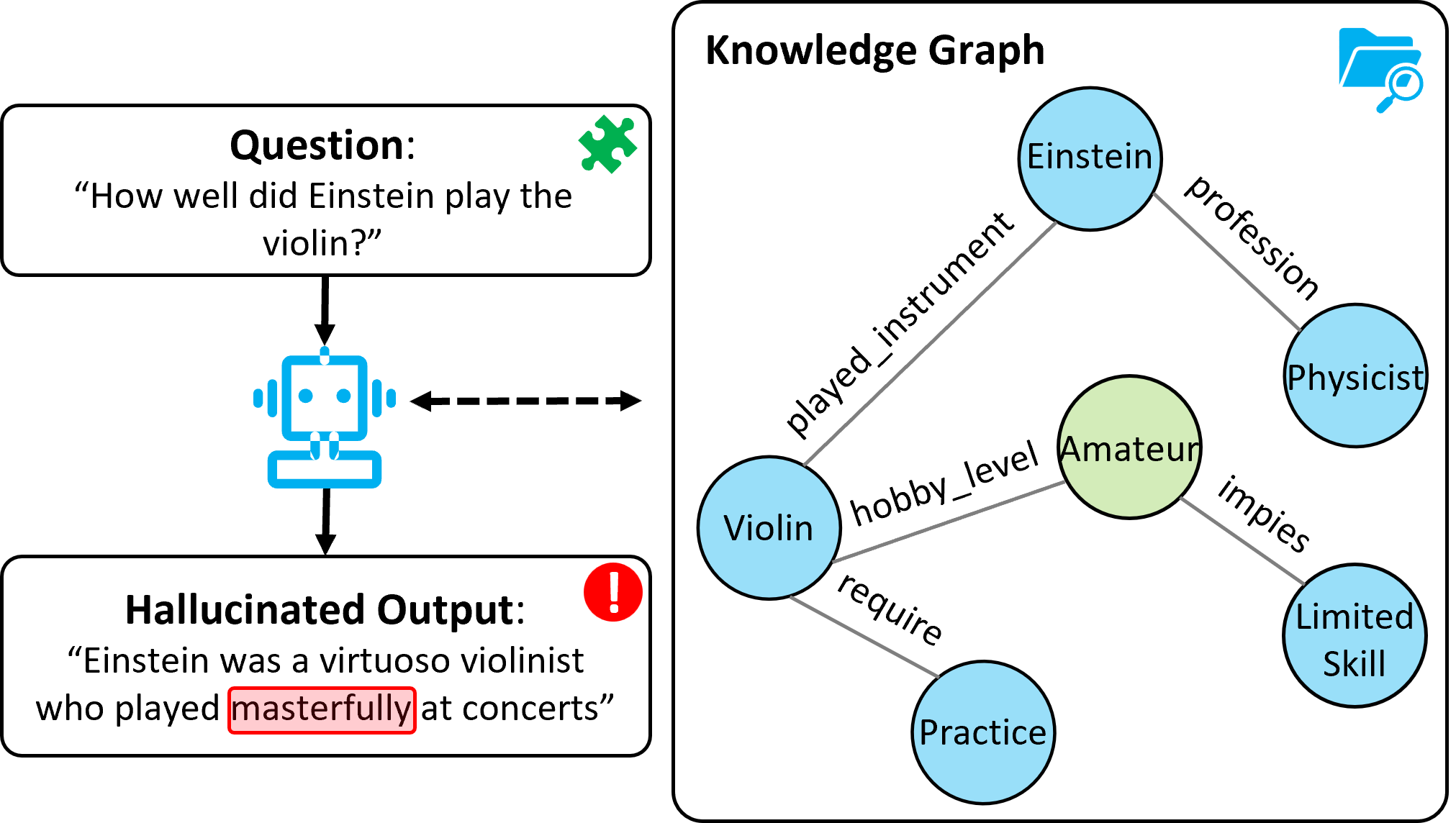}
   \caption{Hallucinations in GraphRAG during Knowledge Base Question Answering (KBQA).}
   \label{fig:hallucinations-example}
\end{figure}

GraphRAG operates in two stages: subgraph retrieval from a knowledge graph, followed by subgraph linearization and answer generation using a language model \cite{han2024retrieval}.
A unique challenge of GraphRAG arises in the subgraph linearization and answer generation stage: the system must convert retrieved graphs into sequences compatible with LLMs, and then rely on the models to interpret and ground this representation. 
To make graph-structured information more accessible to LLMs, prior work has explored various linearization strategies (e.g., adjacency lists, edge-centric formats, triple-based representations) .
Despite these efforts, LLMs still struggle to effectively leverage linearized graph inputs \cite{xypolopoulos2024graph}, due to inherent architectural limitations of the Transformer, which lacks an inductive bias toward relational and topological structures.
These limitations have been shown to contribute to hallucinations in GraphRAG outputs \cite{merrer2024llms} (see Fig.~\ref{fig:hallucinations-example}).

\begin{figure*}[t]
   \centering
   \includegraphics[width=\linewidth]{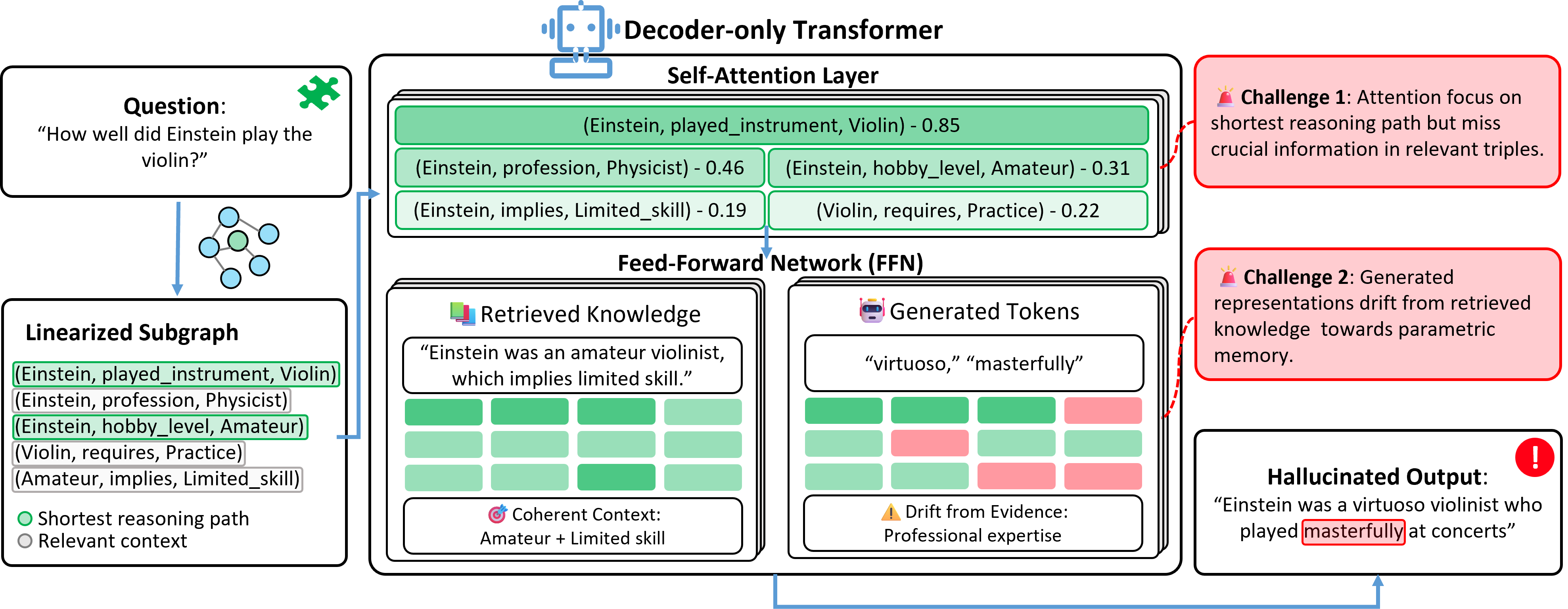}
   \caption{The model over-focuses on shortest paths in the knowledge graph, neglecting relevant context triples (Challenge 1), and the feed-forward layers drift away from retrieved knowledge, relying instead on parametric memory (Challenge 2), leading to hallucinated outputs even when accurate subgraph information is available.}
   \label{fig:hallucinations-reasons}
\end{figure*}

To better understand how this limitation contributes to hallucinations in GraphRAG, we examine the model’s internal processing mechanisms and identify two core challenges related to the attention and feed-forward networks (FFNs) components (See Fig.~\ref{fig:hallucinations-reasons})
First, LLMs often exhibit limited coverage when processing graph-based knowledge. Even when a complete reasoning path is present, attention tends to concentrate on salient or frequently seen triples, neglecting other relevant facts \cite{merrer2024llms}. This shortcut behavior compromises deep reasoning and increases the risk of incomplete or hallucinated answers.
Second, external knowledge remains fragile during the decoding process. Unlike natural language passages that convey information through rich context, semantic flow and contextual redundancy, linearized subgraphs, typically represented as discrete triples (e.g.,``Einstein, played\_instrument, violin”), encode isolated facts with limited coherence or background support. This sparsity hinders the model’s ability to build robust semantic representations in the feed-forward layers, making it more reliant on parametric memory and thus more prone to hallucination \cite{ji2023survey}.

While recent work has proposed new GraphRAG frameworks to enhance performance, a more fundamental question remains underexplored:
\textit{Why do hallucinations persist even when structured knowledge is retrieved and provided to the model?}
Rather than designing a new GraphRAG architecture, this work aims to understand and detect hallucinations by analyzing how LLMs internally process linearized subgraphs.
To this end, we introduce two lightweight interpretability metrics:
(1) \textbf{Path Reliance Degree (PRD)}, which quantifies whether the model disproportionately concentrates attention on shortest-path triples during answer generation, potentially neglecting other relevant information in the linearized subgraph;
(2) \textbf{Semantic Alignment Score (SAS)}, which complements PRD by measuring the degree to which the model’s internal token representations align with the retrieved knowledge triples, thereby capturing failures in semantic grounding.

Through empirical analysis on a knowledge-based question answering (KBQA) dataset, we find that hallucinations are strongly associated with weak semantic grounding, as indicated by low SAS scores. They are further exacerbated when the model either disproportionately concentrates attention on the shortest reasoning path (high PRD), or distributes attention more broadly without successfully grounding the retrieved knowledge.
Based on these findings, we train a lightweight hallucination detector that uses PRD, SAS, and a small set of surface-level features (e.g., unique word ratio, and punctuation counts), which complement internal signals by capturing shallow patterns often associated with hallucinated responses.
Despite its simplicity, the detector achieves strong AUC and F1 performance across two LLMs, outperforming a range of confidence-based and semantic consistency baselines without requiring any labeled answers or model modifications.
By grounding hallucination analysis in mechanistic interpretability \cite{elhage2021mathematical}, our work provides both fundamental insights into LLM limitations with structured knowledge and practical tools for building more reliable GraphRAG systems.

\section{Related Work} 
\subsection{Hallucination in Transformer-Based Language Models}
We focus on decoder-only Transformer architectures, which underlie most state-of-the-art large language models (LLMs) \cite{brown2020language}.
Each layer in such models consists of a masked multi-head self-attention mechanism (Attention Heads), allowing each token to attend to previous tokens, and a feed-forward network (FFN), which applies non-linear transformations independently to each token.
These components contribute differently to knowledge integration during generation. 
Copy heads, a subset of attention heads, inject contextual information directly into the residual stream, helping to preserve and propagate local context across the sequence \cite{olsson2022context}.
FFNs, in contrast, are central to parametric knowledge use in LLMs: they associate frequently observed lexical or semantic input patterns, such as common phrases or topics, with predictive token distributions based on pretraining memory \cite{geva2020transformer}.
Because all input tokens are embedded into a flat sequence and processed uniformly, the model lacks explicit awareness of token origin or structural roles.
As a result, it relies entirely on attention mechanisms to identify relevant context.
This limitation reduces the model’s ability to utilize structured inputs such as retrieved graphs and increases the likelihood of hallucinations, where generation is driven more by internal associations than grounded structure.


\subsection{Limitations of Subgraph Linearization in GraphRAG}
Recent studies show that LLMs struggle to effectively leverage graph-structured information when it is provided in prompt form. 
Although such information can be linearized into prompts, models often rely on token-level semantic associations rather than relational structure.
Experiments reveal that they fail to consistently exploit topological features, such as node adjacency or multi-hop dependencies, even when such structure is explicitly encoded in the input \cite{huang2023can}.
Even with more sophisticated linearization strategies, such as path enumeration, which explicitly lists reasoning paths or traversal-based methods (e.g., breadth-first search), LLMs still struggle to reconstruct global topology and perform multi-hop reasoning effectively \cite{xypolopoulos2024graph}.

While linearizing the retrieved subgraph into a flat text sequence is the dominant approach in GraphRAG \cite{peng2024graph}, prior studies have shown that language models often struggle to effectively utilize such inputs, as they tend to overlook relational and topological information. However, existing work rarely examines how this limitation contributes to hallucination. We address this gap by proposing two lightweight interpretability metrics that reveal how LLMs attend to and retain structured subgraph information during generation.

\subsection{Hallucination Detection via Mechanistic Interpretability}



While RAG systems aim to reduce hallucinations by grounding LLMs in external knowledge, hallucinations still persist.
Recent studies have explored internal mechanisms of LLMs to explain when and why hallucinations occur.
ReDeeP \cite{sun2024redeep} investigates hallucinations from a mechanistic interpretability perspective, showing that attention heads primarily process external context while FFNs encode parametric memory—knowledge encoded in the model weights from pretraining. Their findings suggest that hallucinations often arise when models over-rely on parametric memory encoded in FFNs, especially when attention heads fail to effectively incorporate external context. SEReDeEP \cite{wang2025seredeep} builds on this by introducing semantic entropy to measure model uncertainty and representation drift. It finds that hallucinations often stem not from retrieval failure, where the internal token representations diverge from the semantics encoded in the retrieved knowledge. 

These studies highlight the value of mechanistic interpretability for understanding hallucinations. However, they primarily focus on text-based RAG and overlook how models process structured graph inputs. We extend this line of work by proposing two lightweight interpretability metrics—PRD and SAS—to examine attention concentration and semantic alignment in GraphRAG settings. 
While these methods offer insights into hallucinations in flat, text-based retrieval, they do not address the structural complexity introduced by graph-formatted inputs.
We close this gap by applying interpretability tools to GraphRAG, where hallucinations may result not only from content drift, but also from a failure to preserve graph topology in attention and representation.
\enlargethispage{-\baselineskip} 

\section{Problem Setup and Research Questions}
\subsection{Task Definition}
We study hallucinations in graph-based retrieval-augmented generation (GraphRAG) for knowledge-based question answering (KBQA). In this setting, the model receives:
\begin{itemize}[topsep=3pt, partopsep=0pt, itemsep=2pt, parsep=0pt]
    \item a natural language question $q$;
    \item a subgraph $G = \{(h_i, r_i, t_i)\}_{i=1}^{N}$ composed of entity–relation triples retrieved from a knowledge graph.
\end{itemize}

The objective is to generate a textual answer $\hat{a}$ that correctly answers $q$ using information from $G$. To make the subgraph compatible with large language models (LLMs), we linearize $G$ into a flat sequence of text-formatted triples using a traversal-based serialization strategy \cite{xypolopoulos2024graph}. These triples are then incorporated into a fixed prompt template alongside the question \cite{li2024subgraphrag,brown2020language}. We adopt SQuAD-style evaluation \cite{seo2016bidirectional} to label model outputs as either hallucinated or truthful, which serves as the basis for our further analysis.

While prior work often treats hallucination as a black-box output error, we instead seek to understand how internal behaviors contribute to it. Inspired by recent work that quantitatively measures LLMs’ internal processing dynamics \cite{li2023deceptive,tao2024llms, sun2024redeep, ni2024llms, ni2025towards}, we introduce two complementary interpretability metrics to probe the model's internal behavior.
We focus on KBQA tasks where subgraphs contain clearly defined reasoning paths and ground-truth knowledge, enabling precise measurement of model behavior.

\subsection{Research Questions}
This paper explores a fundamental question in knowledge-grounded generation: \textit{why do LLMs hallucinate even when provided with relevant structured knowledge?} By examining internal model behaviors—specifically attention patterns and feed-forward network (FFN) activations—we aim to uncover the mechanistic basis of hallucination behavior in GraphRAG and to develop lightweight, interpretable detection methods. This controlled experimental setting allows us to establish clear relationships between internal processing patterns and output reliability. We focus on the following research questions:
\begin{itemize}
\item \textbf{RQ1 (Mechanistic Analysis):} \textit{How do the proposed metrics, PRD and SAS, correlate with the presence of hallucinations?}  
We examine whether PRD and SAS capture distinct internal failure modes that reliably differentiate truthful output from hallucinated ones.
\item \textbf{RQ2 (Hallucination Detection):} \textit{Can PRD and SAS support effective hallucination detection?}  
We train and evaluate a lightweight detector using PRD and SAS, demonstrating that it outperforms interpretable baselines based on model confidence and semantic consistency.
\end{itemize}

\section{Interpretability Metrics}
\subsection{Path Reliance Degree (PRD)}
\label{sec:prd}
Inspired by the findings of deceptive semantic shortcuts \cite{li2023deceptive}, we propose the Path Reliance Degree (PRD) to quantify \emph{how strongly a model concentrates attention on a single reasoning path during answer generation}. 
Deceptive semantic shortcuts refer to instances where hallucinations occur despite the availability of the complete gold reasoning chain. This happens when large language models focus on familiar or statistically salient parts of the input, such as frequently seen entities or relations, instead of reasoning through the entire chain. As a result, the model may rely on partial or superficial cues, skipping essential reasoning steps. This often leads to responses that are fluent in form but inaccurate in fact.
To capture this behavior, PRD measures the difference in attention allocated to shortest path tokens versus all other tokens during answer decoding. A disproportionately high PRD suggests that the model may be over-relying on a shortcut-like paths rather than engaging in robust, distributed reasoning, thereby indicating an increased risk of hallucination.

\paragraph{Shortest path tokens.}
We extract \emph{shortest path triples} as all KG triples lying on the shortest paths that connect question‐side entities to answer‐side entities.  
After linearization, each sub-token representing the head, relation, or tail of a shortest path triple is included in the shortest path position set $S$; its complement is denoted a $\bar S$.  
These tokens represent the minimal path connecting the question to the answer.

\paragraph{Formulation.}
Let $A=\{a_1,\dots,a_n\}$ denote the positions of answer tokens—concretely, the tokens that appear after the special prefix ``\texttt{ans:}'' in the decoder output.  
For each decoder layer $l\in[1,L]$ and attention head $h\in[1,H]$, let $a_{l,h,i,j}$ be the raw attention score from answer position $i\!\in\!A$ to source position $j$.  
We normalize with a softmax
\[
\alpha_{l,h,i,j} \;=\;
    \frac{\exp(a_{l,h,i,j})}
         {\sum_{k}\exp(a_{l,h,i,k})}.
\]
The attention mass flowing from $(l,h,i)$ to the shortest path and to non-path tokens is
\[
\alpha_{l,h,i,S}= \sum_{j\in S}\alpha_{l,h,i,j},\qquad
\alpha_{l,h,i,\bar S}= \sum_{j\in\bar S}\alpha_{l,h,i,j}.
\]
Averaging their difference yields
\[
\text{PRD}\;=\;
\frac{1}{L\,H\,|A|}
\sum_{l=1}^{L}\sum_{h=1}^{H}\sum_{i\in A}
\bigl(\alpha_{l,h,i,S}-\alpha_{l,h,i,\bar S}\bigr).
\]
Since each $\alpha_{l,h,i,S}$ within the range $[0,1]$, PRD is theoretically bounded in $[-1,1]$.

\paragraph{Interpretation.}
A \emph{high} PRD indicates that, on average, the decoder assigns a disproportionately large share of attention to shortest path tokens, evidencing \emph{path over-reliance}.  
Our empirical experiments show that such rigid focus frequently co-occurs with hallucinations.  
Conversely, a \emph{low} (or negative) PRD signifies a more distributed allocation of attention across both path and non-path evidence, aligning with flexible semantic reasoning and greater factual accuracy.

\subsection{Semantic Alignment Score (SAS)}
\label{sec:SAS}
While PRD captures the model’s attentional concentration on shortest paths, it does not indicate whether the semantic content of the attended tokens is effectively encoded into the model’s internal representations. Prior research has shown that even when attention is directed toward relevant inputs, the feed-forward layers may still fail to integrate this information into the final representations used for generation \cite{sun2024redeep}. To address this, we introduce the Semantic Alignment Score (SAS), inspired by retrieval matching methods that compare query and document embeddings~\cite{karpukhin2020dense, xiong2020approximate}. Rather than evaluating relevance at the input level, SAS measures the fidelity of knowledge integration by computing the similarity between the model’s internal token-level answer representations and embeddings of the retrieved knowledge. This allows us to assess whether the generated content is semantically grounded in the provided subgraph.

\paragraph{Target expansion set (TES)} 
Starting from the shortest path triples, we expand by selecting neighboring triples that connect to any entity on the path, filtered by semantic relevance criteria based on relation type filtering and connectivity constraints to retain the most informative candidates, yielding the \emph{Target Expansion Set} $\text{TES}(q)$. This expansion enriches semantic content—since shortest path triples are often too sparse for robust embedding comparison—while preserving the essential reasoning structure. 

\paragraph{Token–triple similarity.}
For every answer token $y_t$, we obtain its hidden vector $\mathbf{h}_t$ from a top transformer layer just before the final output
projection, chosen to preserve semantic information while avoiding distortion from generation-specific transformations. Each triple $T_i =  (h_i, r_i, t_i) \in \text{TES}(q)$ is linearized as ``\texttt{h\_i r\_i t\_i}'' and encoded by the same language model (LM); we compute the masked mean of its hidden states to obtain the embedding $\mathbf{g}_i$, where masking excludes padding tokens from the averaging operation. After $\ell_2$ normalization, we define:
\[
 \text{SAS}(y_t) = \max_{T_i \in \text{TES}(q)} \cos\left( \mathbf{h}_t,\, \mathbf{g}_i \right)
\]

\paragraph{Sentence-level score.}
The overall score averages across answer tokens:
\[
    \text{SAS}\;=\;\frac{1}{T}\sum_{t=1}^{T}\text{SAS}(y_t).
\]
More formally, let $\mathbf{h}_t^{(l)} \in \mathbb{R}^d$ denote the $d$-dimensional hidden representation of answer token $y_t$ at layer $l$, and let $\mathbf{g}_i^{(l)} \in \mathbb{R}^d$ be the corresponding encoding of triple $T_i$. After $\ell_2$ normalization:
\[
\text{SAS}(y_t) = \max_{T_i \in \text{TES}(q)} 
\frac{\mathbf{h}_t^{(l)} \cdot \mathbf{g}_i^{(l)}}
{\|\mathbf{h}_t^{(l)}\| \|\mathbf{g}_i^{(l)}\|}
\]
Since cosine similarity is bounded in $[-1, 1]$ and we take the maximum over a finite set, SAS is theoretically bounded: $\text{SAS} \in [-1, 1]$. However, in practice we constrain SAS to $[0, 1]$ to focus on positive semantic alignment, where values closer to 1 indicate stronger grounding in the target knowledge.

\paragraph{Interpretation.}
A larger value indicates stronger semantic grounding in the retrieved subgraph; a smaller value suggests possible drift toward parametric memory and, therefore, higher hallucination risk.

\section{Experimental Setup}
\label{sec:setup}
\paragraph{Task and Dataset.}
We evaluate on MetaQA-1hop \cite{zhang2018variational}, a knowledge graph QA benchmark in the movie domain with 96,106 training, 9,992 development, and 9,947 test questions. 
Each question requires single-hop reasoning, providing a controlled setting for mechanistic analysis where we can precisely define reasoning paths and measure attention allocation.
Following established practice in mechanistic analysis \cite{sun2024redeep, wang2025seredeep}, we use pre-retrieved subgraphs to isolate knowledge processing from retrieval effects. 
We build on the processed dataset from \citet{He-WSDM-2021} with high-quality subgraphs from GraftNet-style pipelines \cite{sun2018open}, which we sequence prune based on shortest paths and target expansion sets to create subgraphs of appropriate length for LLM reasoning. 
This setup ensures that observed hallucinations reflect how LLMs process structured knowledge independent of retrieval quality variations.
For RQ1 empirical analysis, we use 5,000 examples from the development set to examine correlations between PRD, SAS, and hallucination patterns. For RQ2 detector training, we use 37.5k training examples (split into 80\% training and 20\% validation) and evaluate on the full test set. This controlled setup allows us to focus specifically on how LLMs process structured knowledge, independent of retrieval quality variations.

\paragraph{Subgraph Pruning and Expansion.}
To prepare model inputs and analyze grounding behavior, we preprocess the subgraph of each question $G$ using a two-step method. 
First, we apply shortest-path pruning to collect all triples on the shortest paths between gold answer entities and question entities. These triples form the essential reasoning backbone and are guaranteed inclusion in the final subgraph.
Second, we construct a Target Expansion Set (TES) by scoring neighboring triples based on their connectivity to: path entities (3 points), question entities (2 points), and answer entities (2 points). We rank all candidate triples by these connectivity scores.
Finally, we assemble the pruned subgraph by: (1) including all shortest path triples without truncation, then (2) adding the highest-scoring TES triples in rank order until reaching a total of $K$ triples. 
This approach balances essential reasoning chains with relevant context for semantic alignment analysis. For MetaQA-1hop, we set $K=20$ triples, which provides sufficient context while remaining within typical LLM input length constraints. (see appendix~\ref{appedix:pruning} for full pruning process).

\begin{figure}[h]
   \centering
   \setlength{\abovecaptionskip}{3pt} 
   \setlength{\belowcaptionskip}{-3pt} 
   \includegraphics[width=\linewidth]{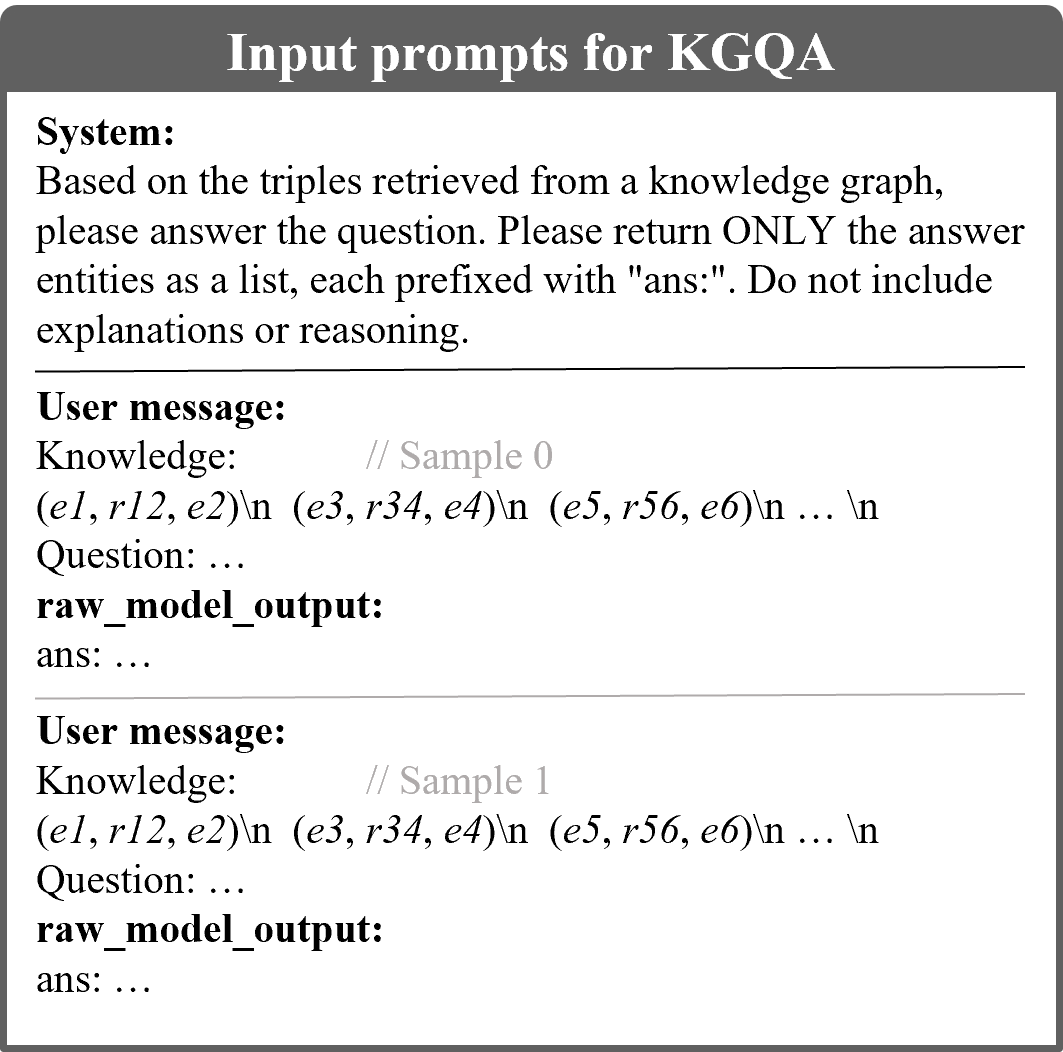}
   \caption{Prompt template for GraphRAG question answering showing linearized knowledge graph triples and answer format requirements.}
   \label{fig:prompt_template_figure}
\end{figure}

\paragraph{Model and Prompting Strategy.}
We use frozen decoder-only language models as base generators: \texttt{Llama-2-7b-chat-hf} \cite{touvron2023llama} for our empirical analysis and both \texttt{Llama-2-7b-chat-hf} and \texttt{Qwen2.5\allowbreak-7B\allowbreak-Instruct} \cite{bai2023qwen} for the building the GGA hallucination detector. 
For each question, we serialize the pruned subgraph into subject–predicate–object triples (e.g., ``Conspiracy release\_year 2001'') and insert them into a fixed prompt template, followed by the natural language question. 
The prompt instructs the model to return only answer entities prefixed with ``ans:'' without explanations or reasoning (As shown in Figure~\ref{fig:prompt_template_figure}).

\paragraph{Hallucination Labeling.} 
\label{sec:lallucination_label}
We adopt SQuAD evaluation metrics \cite{seo2016bidirectional} for hallucination detection, providing more nuanced assessment than simple exact matching alone.
Following standard practice in reading comprehension, we extract and normalize answer entities from model outputs using pattern-based cleaning (removing articles, punctuation, and descriptive patterns), then apply both Exact Match (EM) and token-overlap-based F1 evaluation against gold answers.
A generation is labeled as truthful if it achieves either exact match or sufficient token overlap ($F1 \geq 0.3$), accounting for formatting differences and partial correctness in natural language generation. This refined binary labeling enables accurate analysis of internal model behavior through our interpretability metrics.

\section{RQ1: Empirical Analysis of PRD and SAS}
To answer RQ1, in this section, we conduct three analyses on 5,000 GraphRAG-generated responses to understand how PRD and SAS relate to hallucination in GraphRAG outputs: (1) we compare the distributions of PRD and SAS between hallucinated and truthful responses and assess statistical significance; (2) we examine their joint behavior and correlation to evaluate their complementarity; and (3) we perform a quadrant-based case study to explore hallucination patterns under different PRD–SAS combinations.

\subsection{Distributional Differences and Statistical Significance}
\begin{figure}[H]
   \centering
   \vspace{-.15in}
   \includegraphics[width=\linewidth]{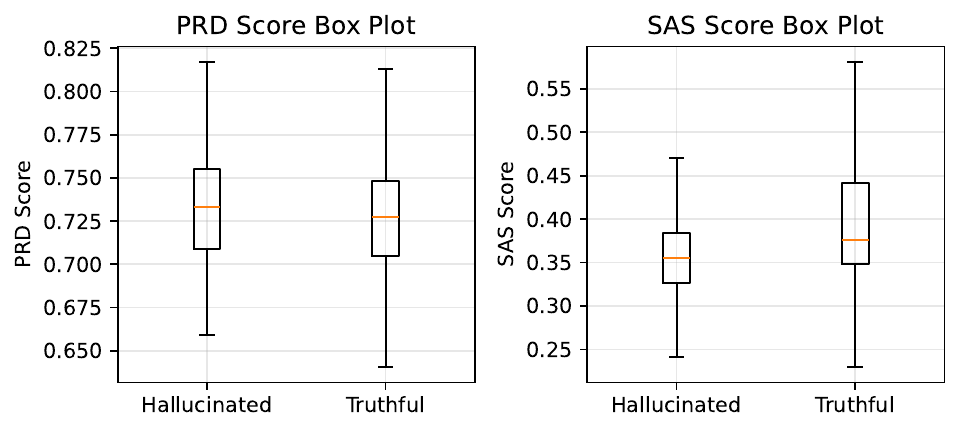}
   \caption{Box plots of PRD and SAS scores for hallucinated vs. truthful responses. Hallucinated answers show higher PRD and lower SAS, indicating stronger shortest path reliance and weaker semantic grounding.}
   \label{fig:boxplots}
   \vspace{-.15in}
\end{figure}

We compare the distributions of PRD and SAS between hallucinated and truthful outputs. As shown in Fig.~\ref{fig:boxplots}, both metrics exhibit statistically significant differences across the two classes.

Two-sample t-tests reveal statistically significant differences between hallucinated and truthful responses, with $t(4998) = -3.31$, $p = 9.48 \times 10^{-4}$ for PRD and $t(4998) = 10.96$, $p = 1.23 \times 10^{-27}$ for SAS. The effect size is small for PRD ($d = -0.18$) and moderate for SAS ($d = 0.60$). These results suggest that hallucinated responses tend to exhibit weaker semantic grounding (lower SAS) and slightly greater reliance on shortest-path triples (higher PRD).

\subsection{Joint Feature Behavior and Complementarity}
\begin{figure}[h]
   \centering
   \setlength{\abovecaptionskip}{2pt} 
   \setlength{\belowcaptionskip}{-3pt} 
   \includegraphics[width=\linewidth]{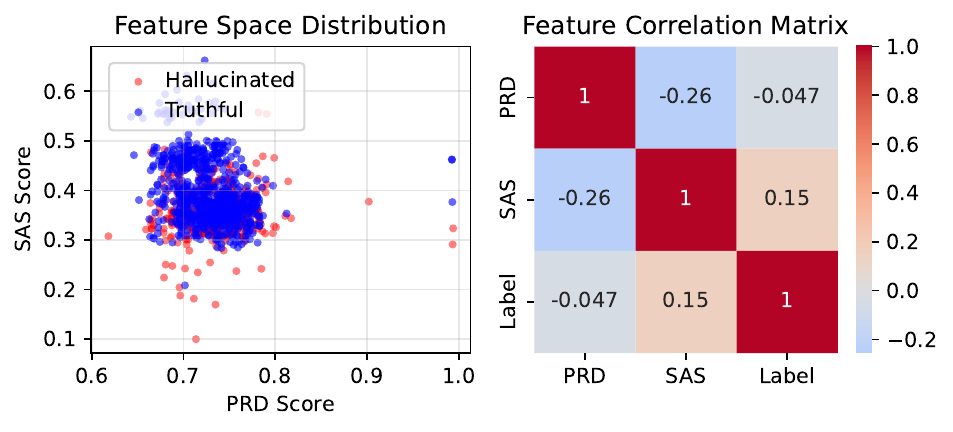}
   \caption{ Feature space and correlation. (\textit{Left}) Scatter plot of TUS vs. SAS shows overlapping but distinguishable clusters. (\textit{Right}) Correlation matrix shows weak dependencies between features and label, suggesting complementary information.}
   \label{fig:feature_analysis}
\end{figure}

We further examine the joint distribution and correlation of PRD and SAS (Fig.~\ref{fig:feature_analysis}). The two signals are weakly correlated ($r = -0.26$), and each aligns differently with hallucination labels (PRD–Label: $r = -0.047$, SAS–Label: $r = 0.15$), indicating complementary contributions. Truthful responses tend to lie in the upper-left region (low PRD, high SAS), while hallucinated ones are more dispersed.

\vspace{-0.5\baselineskip} 

Both PRD and SAS exhibit statistically significant differences between hallucinated and truthful responses, with effect sizes indicating a moderate (SAS) and small (PRD) association. Their low Pearson correlation ($r = -0.26$) suggests that they capture distinct aspects of model behavior, making them promising complementary features for downstream hallucination detection models.

\subsection{Quadrant-Based Case Study of Hallucination Patterns}
To examine how the interaction between PRD and SAS influences hallucination risk, we conduct a quadrant-based analysis. We partition all model outputs into four quadrants based on the median PRD (0.727) and SAS (0.374) values to ensure a balanced distribution:

\begin{itemize}
\item \textbf{Q1 (High PRD, High SAS)}: The model heavily relies on the shortest reasoning paths while maintaining semantic alignment.
\item \textbf{Q2 (Low PRD, High SAS)}: Ideal case: low path dependence with strong semantic grounding.
\item \textbf{Q3 (Low PRD, Low SAS)}: The model avoids over reliance on shortest paths, but its broad attention lacks effective semantic grounding.
\item \textbf{Q4 (High PRD, Low SAS)}: Reasoning is shortest path focused and semantically ungrounded.
\end{itemize}

\begin{figure}[h]
\centering
\includegraphics[width=1.0\linewidth]{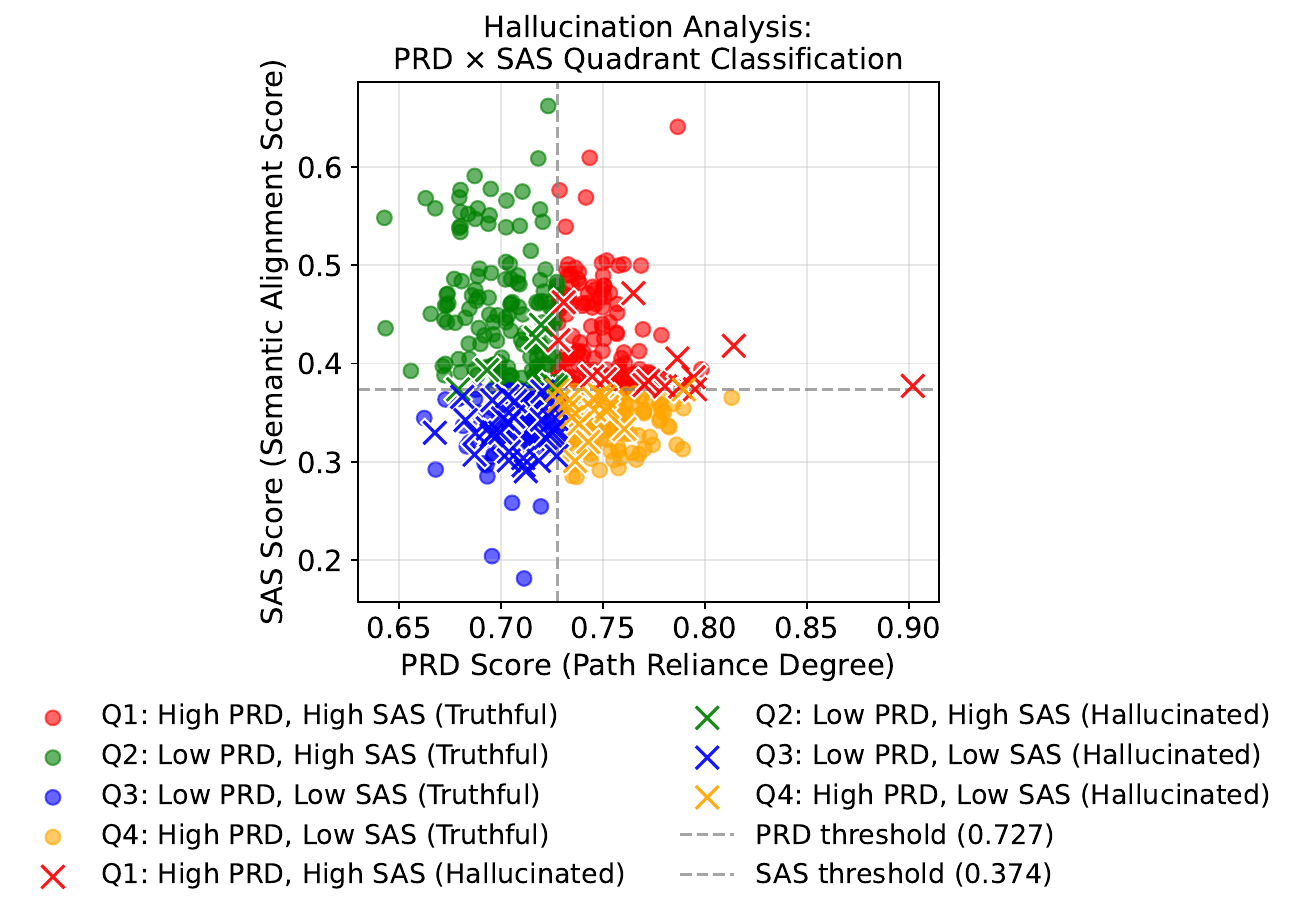}
\caption{Hallucination Analysis via PRD × SAS Quadrant Classification. Each point represents a model output. Hallucinated answers are marked with ``×''.}
\label{fig:quadrant-hallucination}
\end{figure}

We visualize this in Fig.~\ref{fig:quadrant-hallucination} and summarize hallucination rates and average metric values in Table~\ref{tab:quadrant-stats}.

\begin{table}[h]
\centering
\caption{Quadrant-level hallucination rate and average PRD/SAS values. Q3 (low focus, low grounding) shows the highest hallucination rate.}
\small
\begin{tabular}{lcccc}
\toprule
\textbf{Quadrant} & \textbf{Hall.} & \textbf{PRD} & \textbf{SAS} \\
\midrule
Q1: High PRD, High SAS & 9.5\%  & 0.752 & 0.421 \\
Q2: Low PRD, High SAS  & 5.0\%  & 0.701 & 0.452 \\
\rowcolor{gray!15}
Q3: Low PRD, Low SAS   & \textbf{22.2\%} & 0.707 & 0.340 \\
Q4: High PRD, Low SAS  & 10.9\% & 0.754 & 0.344 \\
\bottomrule
\end{tabular}
\label{tab:quadrant-stats}
\end{table}
Interestingly, Q3 (low PRD, low SAS) exhibits the highest hallucination rate (22.2\%), even surpassing Q4 (high PRD, low SAS) at 10.9\%. This counterintuitive finding suggests that distributed attention without semantic grounding is more detrimental than focused attention without grounding. When the model attends broadly across the subgraph but fails to semantically integrate the retrieved information, it may rely on scattered and weakly grounded signals, leading to an even higher risk of hallucination than simple over-reliance on shortest paths.

\section{RQ2: Building a Lightweight Hallucination Detector}
\subsection{Detector Design and Evaluation Setup}
\paragraph{Data and Classifier.}
For detector training, we sample 37.5k examples from the MetaQA-1hop training set, splitting them into 80\% training and 20\% validation. 
We generate answers using frozen LLMs, compute PRD and SAS scores, and apply SQuAD-style exact match to obtain hallucination labels. 
We use XGBoost as our classifier since gradient-boosted trees natively model non-linear feature interactions, handle class imbalance via \texttt{scale\_pos\_weight}, and remain interpretable through feature importance scores.

\paragraph{Feature set.} 
PRD and SAS are our primary mechanism-level features: they capture when the LLM’s attention collapses onto a shortest reasoning path or when its hidden representations drift away from the retrieved graph.  
In addition to these internal reasoning failures, hallucinations often manifest through surface-level textual anomalies.
Prior work has shown that hallucinated responses often exhibit distinctive patterns \cite{ji2023survey}, such as abnormal length, excessive repetition, or formatting inconsistencies. 

To capture these signals without increasing model complexity, we include six \emph{lightweight surface cues}: output length, repetition ratio, average word length, unique-word ratio, an \texttt{ans:} prefix flag, and counts of commas and question marks.  
These statistics are inexpensive to compute, correlate with hallucination risk \cite{manakul2023selfcheckgpt}, and remain fully interpretable.  
Together with PRD and SAS, they provide both an \emph{internal} view of model behavior and an \emph{external} view of output characteristics, improving detection robustness while keeping the feature set compact and transparent.

\paragraph{Baselines} 
We select baseline methods based on two criteria: (1) demonstrated effectiveness in RAG settings and (2) interpretability for fair comparison with our mechanistic approach. 
We evaluate against six widely adopted hallucination detection methods from prior research, falling into two categories: model-internal confidence signals and semantic consistency measures (See Appendix~\ref{appendix:baselines-details} for implementation details).

\textbf{(1) Model-internal confidence metrics} exploit the language model's internal uncertainty signals to detect hallucination. 
This includes \emph{perplexity}, which measures sequence-level likelihood by computing the exponential of cross-entropy loss; \emph{token confidence}, which averages the prediction probabilities across all generated tokens; and \emph{maximum token probability}, which captures the model's highest confidence prediction within the response.

\textbf{(2) Semantic consistency measures} assess the alignment between generated answers and reference knowledge through complementary semantic perspectives. 
\emph{BERTScore} computes contextualized embedding similarity between predicted and gold answers using pre-trained RoBERTa representations \cite{zhang2019bertscore}.
\emph{Embedding divergence} quantifies semantic drift by measuring distributional differences in high-dimensional embedding spaces between generated responses and ground truth, using Jensen-Shannon divergence to compare the probability distributions derived from contextualized embeddings.
\emph{NLI contradiction detection} frames hallucination as logical inconsistency \cite{bowman2015large}, employing natural language inference models to identify contradictions between predictions and reference answers.

\subsection{Results and Analysis}
Table~\ref{tab:baseline_comparison} compares our method (GGA: PRD + SAS) against six hallucination detection baselines across two different LLMs.
Our method consistently outperforms all baselines in terms of AUC and macro-average F1 score, while offering the unique advantage of explaining not only whether but also why responses are likely to be hallucinated.
We report F1 scores both for the hallucinated class (class 1) and macro-averaged across both classes (macro avg), where macro avg computes the unweighted mean of F1 scores across classes to account for class imbalance. For precision and recall, we focus on the hallucination class (class 1), which is most relevant to the task.

Our method consistently outperforms all baselines in terms of AUC and macro-average F1 score. We report F1 scores both for the hallucinated class (class 1) and macro-averaged across both classes (macro avg), where macro avg computes the unweighted mean of F1 scores across classes to account for class imbalance. For precision and recall, we focus on the hallucination class (class 1), which is most relevant to the task.

GGA achieves an AUC of \textbf{0.8341}, a class-1 F1 of \textbf{0.5390}, and a macro F1 of \textbf{0.7524} on \texttt{LLaMA2-7B}, substantially outperforming the strongest semantic baseline, Embedding Divergence (AUC: \underline{0.6914}, class-1 F1: 0.2955, macro F1: \underline{0.5094}). 
Similarly, on \texttt{Qwen2.5-7B}, GGA achieves an AUC of \textbf{0.8528}, a class-1 F1 of \textbf{0.4606}, and a macro F1 of \textbf{0.7083}, again outperforming the best-performing baseline, NLI Contradiction (AUC: \underline{0.5044}, class-1 F1: 0.1150, macro F1: 0.2625).

Notably, while some baseline methods such as Perplexity and BERTScore achieve high recall (e.g., \textbf{0.9310} and \underline{0.9355} on \texttt{LLaMA2-7B} and \texttt{Qwen2.5-7B}, respectively), they suffer from low precision (Perplexity: 0.1484 on \texttt{LLaMA2-7B}; BERTScore: 0.0602 on \texttt{Qwen2.5-7B}), leading to poor class-1 F1 scores (Perplexity: 0.2560; BERTScore: 0.1115) and indicating a tendency to over-predict hallucinations. 
In contrast, our method maintains a better balance between precision and recall (e.g., precision: \textbf{0.5632}, recall: 0.5168 on \texttt{LLaMA2-7B}), while offering clear interpretability that explains why hallucinations occur via attention allocation (PRD) and token-level semantic grounding (SAS).

\begin{table*}[t]
\centering
\setlength{\abovecaptionskip}{4pt}  
\caption{Performance comparisons between our proposed method (GGA: PRD + SAS) and baseline hallucination detectors across two LLMs, grouped by feature category. The boldface represents the best performance, and the underline represents the second-best.}
\label{tab:baseline_comparison}
\small 
\begin{tabular}{@{}lllccccc@{}}
\toprule
\textbf{LLM} & \textbf{Category} & \textbf{Method} & \textbf{AUC} 
& \textbf{\begin{tabular}[c]{@{}l@{}}F1\\ (class 1)\end{tabular}} 
& \textbf{\begin{tabular}[c]{@{}l@{}}F1\\ (macro avg)\end{tabular}} 
& \textbf{\begin{tabular}[c]{@{}l@{}}Precision\\ (class 1)\end{tabular}}
& \textbf{\begin{tabular}[c]{@{}l@{}}Recall\\ (class 1)\end{tabular}} \\
\midrule
\multirow{8}{*}{LLaMA2-7B}
& \multirow{3}{*}{Model-Internal Confidence}
& Perplexity                & 0.4578 & 0.2560 & 0.2126 &  0.1484 &  \textbf{0.9310} \\
&   & Token confidence          & 0.5020 & 0.2020 & 0.4529 & 0.1324 &  0.4249 \\
&   & Max token probability     & 0.5000 & \underline{0.2960} & 0.4478 & 0.1322 & 0.4404 \\
& \multirow{3}{*}{Semantic Consistency}
& BERTScore      & 0.6366 & 0.2740 & 0.5092 & 0.1820 & 0.5544 \\
&   & Embedding Divergence      & \underline{0.6914} & 0.2955 & \underline{0.5094} &  \underline{0.1914} & \underline{0.6476} \\
&   & NLI Contradiction      & 0.5741 & 0.2567 & 0.3524 &  0.1512 & 0.3667 \\
& Our Method               & GGA (PRD + SAS)   & \textbf{0.8341} & \textbf{0.5390} & \textbf{0.7524} & \textbf{0.5632} & 0.5168 \\
\midrule
\multirow{8}{*}{Qwen2.5-7B}
& \multirow{3}{*}{Model-Internal Confidence}
& Perplexity                & 0.4935  & 0.1086  & 0.3512  & 0.0602  & 0.5484 \\
&   & Token confidence          & 0.4966  & 0.1084  & \underline{0.3677}  & 0.0606  & 0.5161 \\
&   & Max token probability     & 0.4419  & \underline{0.1297}  & 0.1904  & \underline{0.0710}  & 0.7500 \\
& \multirow{2}{*}{Semantic Consistency}
& BERTScore       & 0.4867  & 0.1115  & 0.0745  & 0.0593 & \underline{0.9355} \\
&   & Embedding Divergence      & 0.4900  & 0.1134  & 0.0588  & 0.0602  & \textbf{0.9677} \\
&   & NLI Contradiction      & \underline{0.5044} & 0.1150 & 0.2625 &  0.0623 & 0.7419 \\
& Our Method               & GGA (PRD + SAS)      & \textbf{0.8528} & \textbf{0.4606} & \textbf{0.7083} & \textbf{0.3974} & 0.5477 \\
\bottomrule
\end{tabular}
\end{table*}

Overall, GGA consistently achieves the AUC and highest F1 scores on both the hallucinated class and macro average across models, demonstrating its advantage in detecting hallucinations while maintaining balanced and interpretable predictions.
\enlargethispage{-\baselineskip} 

\subsection{Feature Ablation Study}
To evaluate the contribution of different feature groups, we conduct an ablation study, comparing the performance of detectors built on individual and combined subsets of features:

\begin{itemize}[topsep=3pt, partopsep=0pt, itemsep=2pt, parsep=0pt]
\item \textbf{SAS-only}: Uses only semantic alignment signals derived from token–triple similarity.
\item \textbf{PRD-only}: Uses only attention-based signals that capture reliance on shortcut reasoning paths.
\item \textbf{GGA-Core}: Combination of PRD and SAS (mechanistic features).
\item \textbf{GGA-Full}: Mechanistic features plus surface-level cues (e.g., counts of commas and question marks).
\end{itemize}

As shown in Table~\ref{tab:ablation} and Fig.~\ref{fig:ablation}, SAS-only significantly outperforms PRD-only in AUC on both models (0.6324 vs. 0.5502 on LLaMA2-7B, a +15\% improvement; 0.6331 vs. 0.5067 on Qwen2.5-7B, a +25\% improvement), highlighting its strength in identifying hallucinations through semantic grounding. In terms of recall, SAS-only performs better on LLaMA2-7B (0.7291 vs. 0.4763), whereas PRD-only yields slightly higher recall on Qwen2.5-7B (0.6770 vs. 0.6527), indicating variation in attention signal utility across models.

Combining PRD and SAS (GGA-Core) further improves performance, achieving higher AUC than either feature alone (0.6574 on \texttt{LLaMA2-7B} and 0.6730 on \texttt{Qwen2.5-7B}). GGA-Full achieves the best overall AUC (0.8328 on \texttt{LLaMA2-7B} and 0.8506 on \texttt{Qwen2.5-7B}), indicating that surface features enhance classification accuracy. However, GGA-Full’s recall is lower than that of SAS-only (e.g., 0.5223 vs. 0.7291 on \texttt{LLaMA2-7B}), suggesting more conservative predictions that reduce false positives.

Overall, the findings indicate that the foundation of effective hallucination detection relies on the synergy between PRD and SAS, with surface cues helping to refine and enhance prediction precision.

\vspace{-0.2\baselineskip} 
\begin{table}[h]
\centering
\setlength{\abovecaptionskip}{2pt} 
\setlength{\belowcaptionskip}{-3pt} 
\caption{Ablation results across different feature sets. The boldface represents the best performance, and the underline represents the second-best.}
\label{tab:ablation}
\small
\begin{tabular}{@{}l|cc|cc@{}}
\toprule
\textbf{Feature Set} & \multicolumn{2}{c|}{\textbf{LLaMA2-7B}} & \multicolumn{2}{c}{\textbf{Qwen2.5-7B}} \\
 & AUC & Recall & AUC & Recall \\
\midrule
SAS-only   & 0.6324 & \underline{0.7291} &  0.6331 & \underline{0.6527} \\
PRD-only   & 0.5502 & 0.4763 & 0.5067 & \textbf{0.6770} \\
GGA-Core   & \underline{0.6574} & \textbf{0.7374} & \underline{0.6730} & 0.5170 \\
\textbf{GGA-Full} & \textbf{0.8328} & 0.5223 & \textbf{0.8506} & 0.5300 \\
\bottomrule
\end{tabular}
\end{table}

\vspace{-0.8\baselineskip} 
\begin{figure}[h]
\centering
\setlength{\abovecaptionskip}{2pt} 
\setlength{\belowcaptionskip}{-3pt} 
\includegraphics[width=\columnwidth]{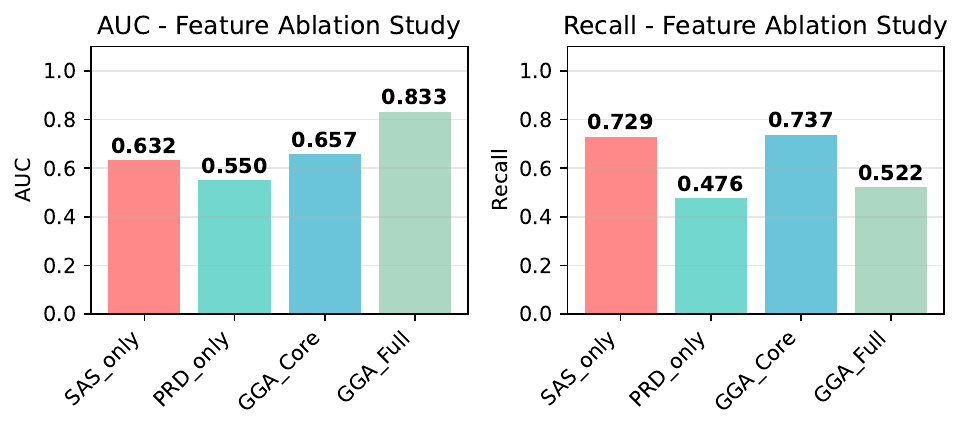}
\caption{Ablation study on LLaMA2-7B showing performance gains from feature combinations. Left: AUC improves as PRD and SAS are combined (GGA-Core), with surface features (GGA-Full) providing further boost. Right: Recall results highlight that mechanistic features effectively capture hallucinations.}
\label{fig:ablation}
\end{figure}

\section{Limitations and Future Work}
We focus on MetaQA-1hop because single-hop questions provide a controlled setting where we can establish clear causal relationships between internal mechanisms and hallucinations. 
This methodological choice enables precise mechanistic analysis by eliminating confounding factors from multi-hop reasoning chains, laying a solid foundation for understanding GraphRAG failures. 
Our findings on attention concentration (PRD) and semantic grounding (SAS) provide generalizable insights that can be extended to more complex reasoning scenarios in future work. 
Additionally, while we compare against established hallucination detection methods, direct comparison with recent end-to-end LLM-based detectors designed for passage-based RAG would be methodologically inappropriate given the structural differences in graph-formatted inputs. 
Our interpretability-based approach offers unique advantages in explaining \emph{why} hallucinations occur, not just detecting them. 
Furthermore, although our experiments focus on decoder-only Transformers, PRD and SAS are designed as architecture-agnostic metrics that capture fundamental attention and representation patterns. 
Validation on encoder-decoder and mixture-of-experts (MoE) models represents a natural extension that could further demonstrate the generalizability of our interpretability framework.
\enlargethispage{-\baselineskip} 

\section{Conclusion}
We identify two internal drivers of hallucination in GraphRAG.  
\emph{Path Reliance Degree} (PRD) reveals that attention collapses onto the shortest reasoning path, whereas \emph{Semantic Alignment Score} (SAS) captures drift between feed‑forward representations and the retrieved subgraph.  
In experiments, PRD and SAS reliably separate hallucinated from truthful answers, and their complementarity enables a plug‑and‑play detector, GGA, that outperforms confidence‑, surface‑, and semantic‑based baselines without any fine‑tuning.  
We anticipate this work will inspire future research on designing GraphRAG systems that leverage LLM internal dynamics as signals, and extending this methodology to multi-hop and open-domain reasoning.

\bibliographystyle{unsrt}
\bibliography{references}

\appendix
\section{Subgraph Preprocessing Details}
\subsection{Subgraph Pruning Algorithm}
\label{appedix:pruning}
Given question entities $E_q$, answer entities $E_a$, and subgraph triples $T$, we construct a pruned subgraph $G'$ of exactly $K$ triples through three stages:
\textbf{Stage 1: Shortest Path Collection.} We identify all shortest paths between question and answer entities using BFS with maximum depth of 3 hops. These path triples form the essential reasoning backbone.
\textbf{Stage 2: Target Expansion Set (TES) Construction.} We collect all 1-hop neighbors of path entities, excluding existing path triples, to form the candidate expansion pool.
\textbf{Stage 3: Connectivity-Based Scoring.} Each TES triple receives a score based on connectivity:
\begin{equation*}
\text{Score}(h,r,t) = 3 \times \text{PathConn} + 2 \times \text{QuestionConn} + 2 \times \text{AnswerConn}
\end{equation*}
Where each connection indicator equals 1 if the triple connects to the respective entity set, 0 otherwise.
\textbf{Final Assembly:} We combine path triples with highest-scoring TES triples until reaching $K$ total triples. 
If insufficient triples exist, we cycle through existing ones to ensure exactly $K$ triples.

\subsection{Scoring Example}
For question \textit{"What genre is the movie Titanic?"} with shortest path Titanic $\rightarrow$ hasGenre $\rightarrow$ Romance:
\begin{itemize}
  \item (Titanic, hasGenre, Romance): \textbf{Score = 5} (path + answer connection)
  \item (Titanic, directedBy, James\_Cameron): \textbf{Score = 5} (path + question connection)
  \item (Romance, isGenreOf, Notebook): \textbf{Score = 2} (answer connection only)
  \item (James\_Cameron, nationality, American): \textbf{Score = 0} (no connections)
\end{itemize}

\subsection{Implementation Notes}
The algorithm runs in $O(|V| + |E|)$ for path finding and $O(|T| \log |T|)$ for scoring. For MetaQA-1hop with $K=20$, path triples typically require 2-4 triples, leaving 16-18 slots for TES expansion. 
This approach balances essential reasoning chains with relevant context for semantic alignment analysis.

\section{Baseline Method Implementation Details}
\label{appendix:baselines-details}
\subsection{Model-Internal Confidence Metrics}
These metrics exploit the language model's internal uncertainty signals to detect hallucination by analyzing the model's confidence in its generated responses.

\subsubsection{Perplexity}
We compute sequence-level perplexity to measure uncertainty over the generated answer:

\begin{equation*}
\text{Perplexity}(y) = \exp\left(-\frac{1}{|y|} \sum_{i=1}^{|y|} \log P(y_i \mid y_{<i})\right)
\end{equation*}

\textbf{Implementation details:}
\begin{itemize}
  \item Use autoregressive loss with \texttt{labels=inputs} for self-supervised computation.
  \item Apply log transformation: $\log(\text{Perplexity})$ for numerical stability.
  \item Clip values to $[1.0, 10000]$ to prevent overflow.
  \item Higher perplexity indicates greater uncertainty and potential hallucination.
\end{itemize}

\subsubsection{Token Confidence}
We calculate the average maximum prediction probability across generated tokens:

\begin{equation*}
\text{TokenConf}(y) = \frac{1}{|y|} \sum_{i=1}^{|y|} \max_v P(v \mid y_{<i})
\end{equation*}

\textbf{Implementation details:}
\begin{itemize}
  \item Apply softmax to logits: $P(v \mid y_{<i}) = \text{softmax}(\text{logits}_i)_v$.
  \item Confidence values are averaged across all token positions.
  \item Values clipped to $[0.2, 0.95]$ to improve robustness.
  \item Lower confidence signals a higher risk of hallucination.
\end{itemize}

\subsubsection{Maximum Token Probability}
We extract the maximum token prediction probability from the response:

\begin{equation*}
\text{MaxTokenProb}(y) = \max_{i=1}^{|y|} \max_v P(v \mid y_{<i})
\end{equation*}

\textbf{Implementation details:}
\begin{itemize}
  \item Use softmax over vocabulary to compute probabilities.
  \item Select maximum probability across all tokens and vocab.
  \item Clip to $[0.4, 0.95]$ to reduce noise.
  \item High values may indicate overconfident hallucinations.
\end{itemize}

\subsection{Semantic Consistency Measures}
These metrics assess alignment between generated answers and the input question from different semantic perspectives.

\subsubsection{BERTScore}

We compute contextual embedding similarity via the official BERTScore implementation:

\begin{equation*}
\text{BERTScore-F1} = \frac{2 \cdot P_{\text{BERT}} \cdot R_{\text{BERT}}}{P_{\text{BERT}} + R_{\text{BERT}}}
\end{equation*}

\textbf{Implementation details:}
\begin{itemize}
  \item Use \texttt{bert\_score.score(candidates, references, lang="en")} with RoBERTa-large.
  \item Fallback: SentenceTransformer \texttt{all-MiniLM-L6-v2}.
  \item Compare generated answers to the question (not ground truth answers).
  \item F1 scores are normalized to the $[0, 1]$ range.
\end{itemize}

\subsubsection{Embedding Divergence}
We compute semantic drift between the question and generated answer embeddings:

\begin{equation*}
\text{EmbedDiv}(q, a) = 0.4 D_{\text{cos}} + 0.2 D_{\text{euc}} + 0.2 D_{\text{angle}} + 0.2 D_{\text{JS}}
\end{equation*}

Where:
\begin{align*}
D_{\text{cos}} &= 1 - \frac{\mathbf{e}_q \cdot \mathbf{e}_a}{\|\mathbf{e}_q\|\|\mathbf{e}_a\|} \\
D_{\text{euc}} &= \min\left( \frac{\|\mathbf{e}_q - \mathbf{e}_a\|}{2\sqrt{d}}, 1 \right) \\
D_{\text{angle}} &= \frac{\arccos(\text{clip}(\cos \theta, -1, 1))}{\pi} \\
D_{\text{JS}} &= \text{JSD}(\text{softmax}(\mathbf{e}_q), \text{softmax}(\mathbf{e}_a))
\end{align*}

\textbf{Implementation details:}
\begin{itemize}
  \item Use \texttt{all-MiniLM-L6-v2} for embedding extraction.
  \item Combine four normalized metrics with empirically-tuned weights.
  \item Higher divergence reflects greater semantic mismatch.
\end{itemize}

\subsubsection{NLI Contradiction Detection}
We detect logical inconsistencies using a pretrained NLI model:

\begin{equation*}
\text{NLIContra}(q, a) = P(\text{contradiction} \mid \text{premise}=q, \text{hypothesis}=a)
\end{equation*}

\textbf{Implementation details:}
\begin{itemize}
  \item Model: \texttt{facebook/bart-large-mnli} via HuggingFace pipeline.
  \item Input format: \texttt{[question] [SEP] [answer]}.
  \item Extract contradiction probability from 3-class output.
  \item Enhance scores: amplify values $> 0.5$, compress $< 0.5$.
  \item Higher scores indicate more likely hallucinations.
\end{itemize}

\subsection{Hyperparameter Settings}
\subsubsection{Model Configuration}
\begin{itemize}
  \item \textbf{Base Models:} \texttt{meta-llama/Llama-2-7b-chat-hf}, \texttt{Qwen2.5\allowbreak-7B\allowbreak-Instruct}
  \item \textbf{Precision:} \texttt{torch.float16} on GPU, \texttt{torch.float32} on CPU
  \item \textbf{Device Map:} \texttt{device\_map="auto"}
  \item \textbf{Memory Limit:} 6GB per model
\end{itemize}

\subsubsection{Processing Parameters}
\begin{itemize}
  \item \textbf{Batch Size:} 8 (RTX 4070), 2 (RTX 3060)
  \item \textbf{Max Length:} 64 tokens
  \item \textbf{Truncation:} Enabled for all tokenization
  \item \textbf{Memory Management:} Call \texttt{torch.cuda.empty\_cache()} every 10–20 samples
\end{itemize}

\subsubsection{Classification Settings}
\begin{itemize}
  \item \textbf{Classifier:} LogisticRegression with \texttt{class\_weight="balanced"}
  \item \textbf{Feature Scaling:} StandardScaler, with clipping at $\pm 3\sigma$
  \item \textbf{Threshold Selection:} Grid search over $[0.1, 0.9]$ (50 steps), optimizing F1
  \item \textbf{Cross-Validation:} 3-fold CV
  \item \textbf{Random State:} 42 (global)
\end{itemize}

\vspace{0.3em}
This setup ensures robust and reproducible baseline evaluations under resource-constrained settings.

\end{document}